\documentclass[letterpaper]{article} 
\usepackage[preprint]{aaai2027}  
\usepackage[hyphens]{url}  
\usepackage{graphicx} 
\usepackage{natbib}  
\usepackage{caption} 
\usepackage{algorithm}
\usepackage{algorithmic}

\usepackage{newfloat}
\usepackage{listings}
\DeclareCaptionStyle{ruled}{labelfont=normalfont,labelsep=colon,strut=off} 
\floatstyle{ruled}
\newfloat{listing}{tb}{lst}{}
\floatname{listing}{Listing}

\usepackage{booktabs}

\title{Double-Rollout Reinforcement Learning with Sampling-based Reward Estimation}
\author{
    Haifeng Qian\equalcontrib,
    David Mosallanezhad\equalcontrib,\\
    Christian Munley,
    Adithya Renduchintala,
    Oleksii Kuchaiev
}
\affiliations{
    NVIDIA\\
    2788 San Tomas Expressway, Santa Clara, CA 95051, USA\\
    \{haifengq,dmosallanezh,cmunley,adithyare,okuchaiev\}@nvidia.com
}

\begin{document}

\maketitle

\begin{abstract}
Reinforcement learning with verifiable rewards (RLVR) implemented with Group Relative Policy Optimization (GRPO) has driven substantial progress in training reasoning language models. However, verifiable outcome rewards provide an imperfect measure of reasoning quality: a correct final answer may not be supported by the reasoning trace, whereas an incorrect rollout may contain a nearly correct reasoning trajectory. We present empirical evidence for this mismatch and hypothesize that the resulting reward noise contributes to instability in reinforcement learning for both reasoning accuracy and token efficiency. To address this credit assignment challenge, this paper proposes \textit{double-rollout reinforcement learning}. For each sampled rollout, we truncate its reasoning trace, sample multiple continuations from the truncated prefix, and replace the original reward with the mean continuation reward, yielding a more robust estimate of reasoning quality. This double-rollout algorithm incurs only a moderate increase in compute cost and yet reaches substantially higher accuracies across math and coding tasks and across Qwen and Nemotron models than baseline algorithms with fewer RL steps. For example, using only DAPO data it trains Nemotron-3-Nano-30B-A3B to reach 95.4\% on AIME 2026, versus 91.5\% by GRPO and substantially above state of the art for this model size, within 30 RL steps. When combined with length-based rewards, the same algorithm further improves the accuracy-efficiency Pareto frontier over existing efficiency optimizations. These findings suggest that the double-rollout reward estimation is an effective and cost-efficient way to improve both the accuracy and token efficiency of reasoning RL.

Training code is available at:\\
\scriptsize{\url{https://github.com/NVIDIA-NeMo/RL/tree/haifengqian/double-rollout-paper}}

\end{abstract}


\section{Introduction}\label{sec:intro}

Reasoning large language models (LLM) trained by reinforcement learning with verifiable rewards (RLVR) have been the foundation of the past few years' tremendous progress in artificial intelligence \cite{jaech2024openai,guo2025deepseek,team2026kimi,zeng2026glm,qwen35blog,blakeman2026nemotron}.
A crucial ingredient of successful RLVR is building reliable verifiers, -- carefully designed rule-based or model-based judges to assign proper reward values and to close loopholes of reward hacking, -- because unreliable rewards lead to unstable and suboptimal RL.

However, one aspect of RL reward reliability remains a great challenge: even perfect verifiers would assign dubious reward values to RL rollouts when an LLM takes a guess based on an inconclusive chain of thought.
We will present an empirical study in Section~\ref{sec:method} to measure the prevalence of this problem on math RL rollouts where the verifiers are nearly perfect.
These incorrect reward values can happen in both directions: the model might guess a correct final answer when the chain of thought does not support it, in which case RL training would force the model to learn this questionable sequence with positive advantage; the model might also generate a wrong final answer when its inconclusive chain of thought is on a correct path, in which case RL training would force the model to push down the probability of this chain of thought.
Both scenarios contribute to unstable and suboptimal RL and they are special cases of the credit assignment challenge in RL.

This is particularly damaging when the RL training includes length-based reward or penalty, attempting to achieve better token efficiency.
As we will show in the empirical study, the chance of these incorrect reward values is inversely correlated with rollout sequence lengths.
In other words, when we assign extra high reward values to a short and correct rollout, this rollout is often correct only by chance and we would be forcing the model to learn from a subpar sequence where most of the tokens in the thinking trace are of low quality.
This explains why existing RL methods for token efficiency often lead to undesirable trade-offs, some of which are included as baselines in Section~\ref{sec:results}. 

One possibility to mitigate this problem of unreliable rewards is to include some form of process reward model (PRM) during RL \cite{zhang2025lessons}, such as Proximal Policy Optimization (PPO)'s value network \cite{schulman2017proximal}.
However, the field has long recognized the challenges of training and using PRMs and, thanks to its simplicity and robustness, Group Relative Policy Optimization (GRPO) and its variants have become the main-stream approach for training reasoning LLMs \cite{shao2024deepseekmath}, instead of PPO or other algorithms that involve PRMs.
For one thing, a PRM would provide more surface for reward hacking and introduce new sources of unreliable rewards.

In this paper, we address the unreliable-reward problem by extending the simplicity and robustness of GRPO, by leveraging the fact that we can estimate the value of a partial sequence by sampling multiple continuations to finish and calculating the average verifiable reward of the continuations.
To control the extra cost of sampling, we pick a practical middle ground: for each rollout, we truncate it at one place and perform the sampling-based reward estimation.
For example, by truncating at the 90\% of each rollout's thinking trace and performing 8 continuations, we approximately double the compute needed for rollouts.
With a simple scheme of replacing the original reward with the sampled estimation, more details in Section~\ref{sec:method}, we observe remarkable improvements in RL.
For example, using only DAPO data\footnote{\url{http://huggingface.co/datasets/BytedTsinghua-SIA/DAPO-Math-17k}} \cite{yu2025dapo} we are able to train Nemotron-3-Nano-30B-A3B\footnote{\url{https://huggingface.co/nvidia/NVIDIA-Nemotron-3-Nano-30B-A3B-BF16}} \cite{blakeman2025nemotron} to reach 95.4\% accuracy on AIME 2026, comparable to models with ten times its size.
We significantly outperform baseline GRPO algorithms across Qwen and Nemotron model families and across math and coding reasoning tasks.
When used in combination with length-based rewards, the proposed method reaches accuracy-efficiency trade-off points across a wide spectrum that are superior to those reachable by GRPO with length-based rewards.
We also observe a reduction in the number of RL steps needed to achieve best accuracy, which offsets some of the per-step compute cost increase.

The proposed method is also applicable to reinforcement learning from human feedback (RLHF) on reasoning LLMs. We'll leave that to future work.

\section{Double-Rollout RL}\label{sec:method}

This section starts with a study on the unreliable-reward problem followed by details of the proposed RL algorithms.

\begin{figure}[t]
\centering
\includegraphics[width=0.98\columnwidth]{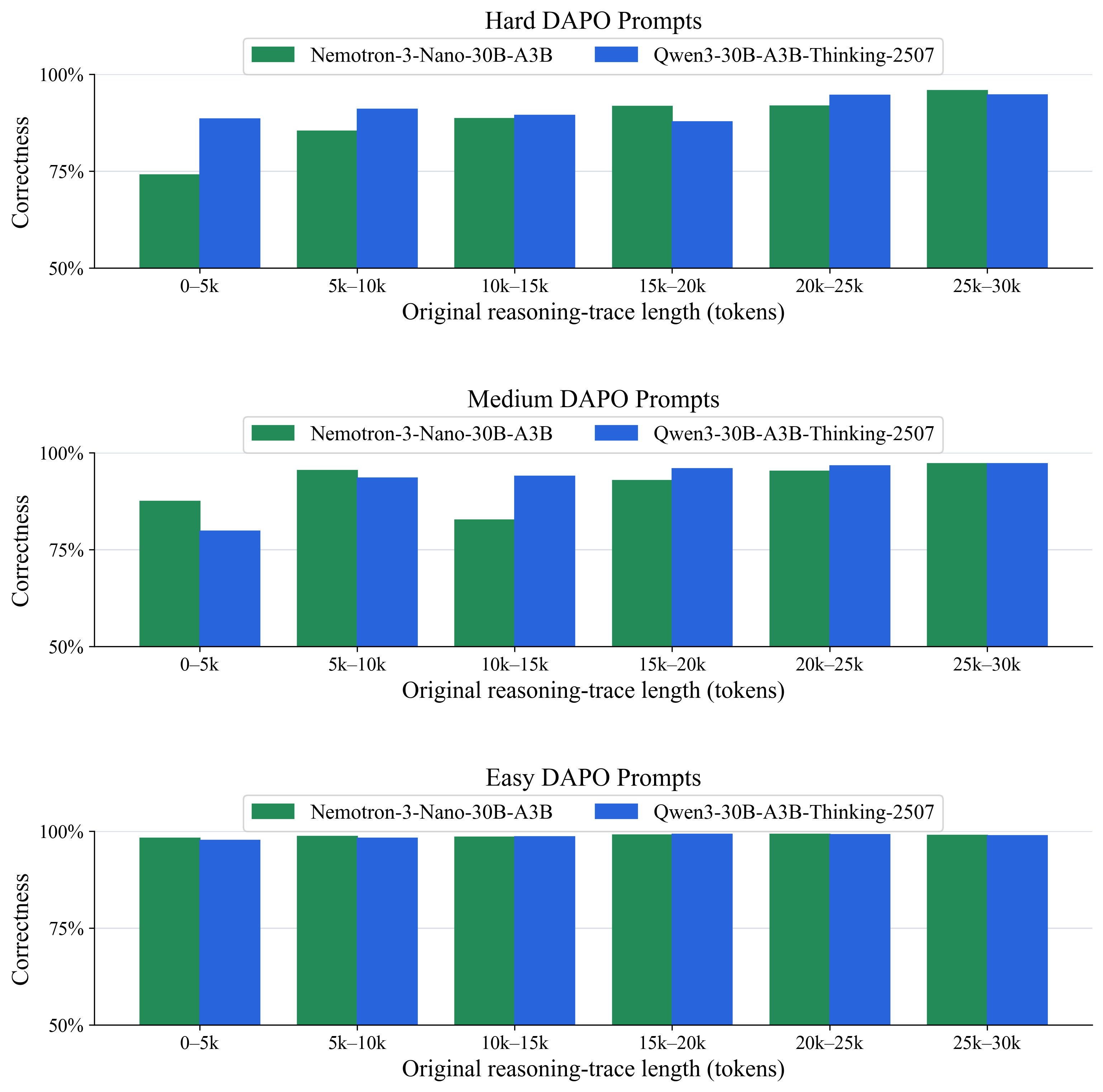}
\caption{Average correctness of a re-generation continuing from 90\% of a {\bf\emph{correct}} rollout's thinking trace. X-axis is the length of the original correct rollout. The prompts are DAPO prompts divided into three subsets by difficulty.}
\label{fig:passrate_posi}
\end{figure}

\begin{figure}[ht]
\centering
\includegraphics[width=0.98\columnwidth]{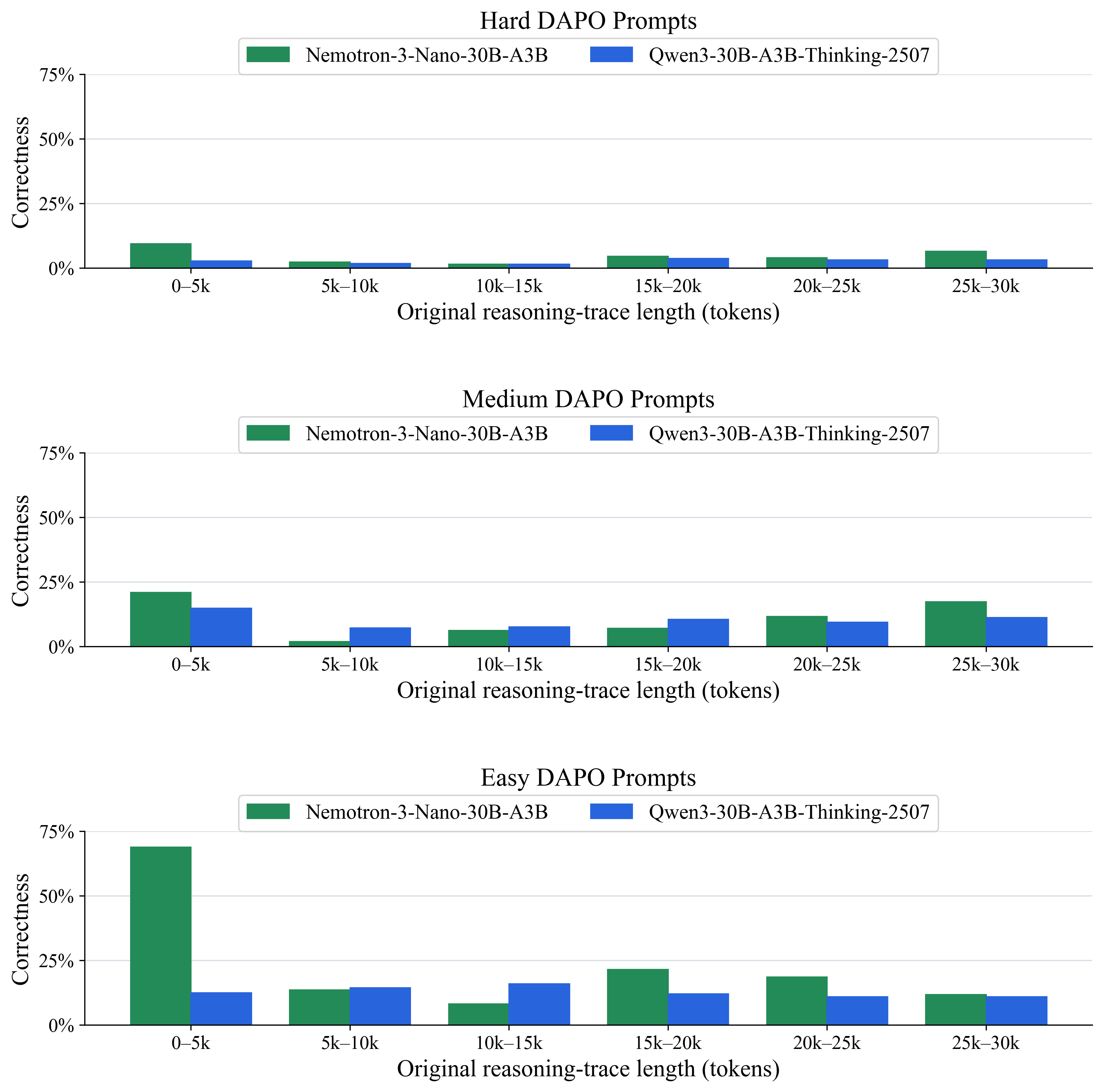}
\caption{Average correctness of a re-generation continuing from 90\% of an {\bf\emph{incorrect}} rollout's thinking trace. X-axis is the length of the original incorrect rollout. The prompts are DAPO prompts divided into three subsets by difficulty.}
\label{fig:passrate_neg}
\end{figure}

\subsection{An Empirical Study on the Reliability of Rewards}

As discussed in Section~\ref{sec:intro}, even perfect verifiers would assign dubious reward values to RL rollouts when an LLM takes a guess based on an inconclusive chain of thought.
We conduct an experiment to quantify this effect and present the results in Figures~\ref{fig:passrate_posi} and \ref{fig:passrate_neg}.

We collect rollouts by Nemotron-3-Nano-30B-A3B and Qwen3-30B-A3B-Thinking-2507 on DAPO prompts.
We truncate each rollout and keep the first 90\% of its thinking trace.
We generate continuations starting from each truncated rollout and measure the average correctness of the continuations.
We use the math verifier in NeMo-RL\footnote{\url{https://github.com/NVIDIA-NeMo/RL}} \cite{nemo-rl} which is nearly perfect on DAPO problems.
We group the results by the correctness of the original rollout, showing correct and incorrect ones in Figures~\ref{fig:passrate_posi} and \ref{fig:passrate_neg} respectively.
Within each figure, we further divide the DAPO problems into three difficulty-level subsets based on pass-rates by the two models, showing as three subplots.
Within each subplot, we further group the data points by binning the lengths of the original rollouts.

It's not surprising that the continuations based on 90\% of a correct rollout are sometimes wrong or those based on an incorrect rollout are sometimes correct.
What's surprising the the magnitude of the change.
As shown by the top subplot of Figure~\ref{fig:passrate_posi}, for a short and correct rollout on a hard DAPO problem, the pass rate of the continuations is on average as low as 74\%.
In other words, when we assign high reward values to these short and correct rollouts during RL, we would force the model to learn low quality sequences, or at least 90\% of these positive-advantage training tokens are of low quality.

The problem also exists in the opposite direction.
Figure~\ref{fig:passrate_neg} suggests that a substantial fraction of incorrect rollouts contain thinking traces that are on the right paths to correct solutions.
Consequently, the probabilities of some good quality reasoning tokens are unfairly pushed down during RL.

A consistent observation across all six plots is the correlation between correctness of continuations and the length of the original rollout.
To be precise, the shorter an original rollout is, the larger the discrepancy is between the original correctness and the average continuation correctness.
We hypothesize that this phenomenon is behind the undesirable trade-offs by existing RL methods to optimize token efficiency \cite{liu2025dler,team2026kimi}: because shorter rollouts' quality are less certain, when we encourage the model to learn from them we have inadvertently encouraged uncertainty.

\subsection{Double-Rollout Algorithm for Max Accuracy}

As discussed in Section~\ref{sec:intro}, we would like to address the unreliable-reward problem without introducing the complexity and instability of PRMs.
Instead, we propose a minimalist method by extending the spirit of GRPO: during RL, we apply the continuation sampling scheme as detailed in the empirical study and simply replace each original reward value with the average reward of the continuations.
Despite its simplicity, this algorithm change leads to significant benefit over vanilla GRPO, to be presented in Section~\ref{sec:results}.
The following are additional engineering details.

After our reward revisions, there exist correctness-vs-reward mismatched rollouts: a rollout with a correct final answer could receive low reward, while a rollout with an incorrect final answer could receive high reward.
The former case is because, during the original rollout, the model makes a correct guess based on an inconclusive or even incorrect thinking trace.
The latter case is because the model makes an incorrect final guess although the thinking trace is towards the right direction.
We observe empirically that fixing the latter case, -- by simply replacing the original rollout with one of the correct continuations, -- helps RL training results, while fixing the former case is unnecessary.
This makes intuitive sense: the former case should be punished despite the correct final answer because we want to discourage premature guessing.
Therefore, the proposed RL algorithm includes one-sided rollout replacements: for each original rollout that receives positive GRPO advantage after our reward revisions and yet contains an incorrect final answer, we replace it with one of its correct continuations.

A major consideration of the proposed RL algorithm is finding a sweet spot between the cost and benefit of the extra sampling incurred during the second rollout stage.
As the example mentioned in Section~\ref{sec:intro}, by truncating at the 90\% of each rollout's thinking trace and performing 8 continuations, we approximately double the overall compute for rollouts.
To avoid costly long tails, we set an upper limit for the number of generated tokens in the second rollout as\footnote{The limit is only used during RL and is not used during the empirical study in the previous subsection.}: $$10\%\cdot\left(\mathrm{original\ thinking\ trace\ length}\right)+\mathrm{buffer}$$
where $\mathrm{buffer}$ is task dependent and we have found empirically 1K to be sufficient for math prompts while 4K to be sufficient for coding.
Intuitively, the second rollout stage gives the model sufficient token budget to recover the original thinking trace plus a task dependent budget for generating responses.
It's worth noting that, because the second rollout stage involves many parallel short generations, it's more friendly for maximizing GPU utilization than the first rollout stage.

\subsection{Double-Rollout Algorithm for Token Efficiency}

As suggested by the empirical study, the proposed sampling-based reward estimation should be able to mitigate the training instability of existing RL methods for token efficiency and achieve better accuracy-efficiency trade-off for reasoning LLMs.
In theory, the proposed method can be used in conjunction with any existing proposals for length-based RL penalty/reward.
In this paper we will focus on modifying \cite{arora2025training} to develop a specific algorithm and leave other variants to future work.

The key idea of \cite{arora2025training} is to use the same group-relative method of GRPO to calculate length-based penalty and thereby avoid introducing absolute-token-count hyper-parameters.
Specifically, it calculates the mean and standard deviation of the lengths of rollouts and a length-based penalty is defined as a function of the z-score of the length of a rollout; the penalty is only applied on correct rollouts.
We adapt this method to work jointly with our double-rollout reward calculation, with the following engineering details:
\begin{itemize}
    \item Instead the subset of correct rollouts, we work on the subset of rollouts with positive GRPO advantage after our reward revisions. Our calculation of the mean and standard deviation of the lengths is also based on this subset.
    \item We apply reward instead of penalty. In others words, we revise the rewards for short-and-reliably-correct rollouts upward, proportionally to its z-score, but we do not apply any length-based downward revision. This is a heuristic choice that allows the freedom to arbitrarily amplify the magnitude of length-based reward without any risk of confusing the model between incorrect rollouts versus correct-but-long rollouts.
\end{itemize}

\section{Experiments}\label{sec:results}

\begin{table*}[t]
\centering
\begin{tabular}{llccc|cc}
  \toprule
  Start model & Algorithm & \multicolumn{3}{c}{RL with DAPO prompts} &
  \multicolumn{2}{c}{RL with OCR prompts} \\
  & & AIME 2024 & AIME 2025 & AIME 2026 & LCB v5 & LCB v6 \\
  \midrule
  & Start & 90.0\% & 88.3\% & 89.6\% & 72.4\% & 67.6\% \\
  Nemotron-3-Nano & GRPO  & 91.7\% & 89.2\% & 91.5\% & 72.5\% & 70.9\% \\
  -30B-A3B & GRPO w/ large rollout & 90.8\% & 89.0\% & 91.9\% & 70.7\% & 70.6\% \\
  & Double-rollout (ours) & 92.1\% & 93.8\% & 95.4\% & 74.6\% & 70.7\% \\
  \midrule
  & Start & 92.5\% & 83.8\% & 87.1\% & 68.3\% & 66.3\% \\
Qwen3-30B-A3B & GRPO & 93.3\% & 88.5\% & 89.0\% & 74.4\% & 70.8\% \\
-Thinking-2507 & GRPO w/ large rollout & 90.4\% & 87.1\% & 87.5\% & 72.2\% & 74.1\% \\
& Double-rollout (ours) & 93.3\% & 90.8\% & 90.0\% & 76.8\% & 72.1\% \\
  \bottomrule
\end{tabular}
\caption{Comparing the double-rollout RL algorithm for max accuracy against RLVR baselines on best achievable accuracies.}
\label{table:accu}
\end{table*}

Table~\ref{table:accu} compares the proposed double-rollout algorithm for max accuracy against baselines.
We conduct experiments in four settings: two choices of starting models times two choices of RL training data.
The two starting models are Nemotron-3-Nano-30B-A3B \cite{blakeman2025nemotron} and Qwen3-30B-A3B-Thinking-2507\footnote{\url{https://huggingface.co/Qwen/Qwen3-30B-A3B-Thinking-2507}} \cite{qwen3technicalreport}, across two different model families.
The two RL training datasets are DAPO \cite{yu2025dapo} for math and OpenCodeReasoning (OCR)\footnote{\url{https://huggingface.co/datasets/nvidia/OpenCodeReasoning-2}; \url{https://huggingface.co/datasets/nvidia/Nemotron-RL-coding-competitive_coding}} \cite{ahmad2025opencodereasoning} for coding, across two different types of reasoning tasks.
Both datasets are publicly available and predate the two starting models.
Based on the entropy of the two starting models' generations, it is highly likely that Qwen3-30B-A3B-Thinking-2507 has been heavily trained on the DAPO data and that Nemotron-3-Nano-30B-A3B has been heavily trained on the OCR data.
Therefore, it is not surprising that training on these prompts again using vanilla GRPO achieves limited improvements on these two models.

For each of the four settings, we compare the proposed algorithm against two baseline algorithms: vanilla GRPO and GRPO with large rollout.
For vanilla GRPO and the double-rollout method, we use 128 prompts per batch and 16 rollouts per prompt.
For the double-rollout method, we truncate at 90\% of thinking traces and use 8 continuations per rollout.
For GRPO with large rollout, we also use 128 prompts per batch but with 128 rollouts per prompt; in other words, the number of rollouts per prompt matches the number of continuations per prompt in the double-rollout runs.
Note that these GRPO-with-large-rollout RL runs are extremely expensive and the per-step cost is 8 times that of the vanilla GRPO.
As discussed in previous sections, the double-rollout scheme approximately double the overall compute for rollouts and hence much cheaper than GRPO with large rollout.
All experiments are performed with NeMo-RL \cite{nemo-rl} and data profiling is done for all experiments to remove prompts with zero or 100\% pass rates.
For each RL run with DAPO data, we pick the checkpoint with the highest sum of AIME 2024, 2025 and 2026 accuracies and report in Table~\ref{table:accu}.
For each RL run with OCR data, we pick the checkpoint with the highest sum of LCB v5 and v6 accuracies and report in Table~\ref{table:accu}.

Across all four settings, the proposed double-rollout RL method reaches the best accuracies.
It is remarkable that it is able to train a 30B MoE model to reach 95.4\% accuracy on AIME 2026 using only outdated training data.
This is an accuracy level that is comparable to models of 10 times its size trained on more and newer data.

For each of the four settings, the number of RL steps for GRPO to reach peak accuracy is 30, 45, 44 and 45 respectively.
In comparison, the number of RL steps for the double-rollout method to reach peak accuracy is 30, 18, 16 and 30.
We hypothesize that, because each double-rollout RL step presents the model with higher quality training data, the RL run becomes more data efficient and requires less steps.
This helps offset some of the extra per-step compute incurred by our proposal.

\begin{table}[t]
\centering
\begin{tabular}{lccc}
  \toprule
  Param  & \multicolumn{3}{c}{RL with DAPO prompts} \\
  & AIME 2024 & AIME 2025 & AIME 2026 \\
  \midrule
  0-25\% & \multicolumn{3}{c}{degradation from start model}\\
  50\% & 93.5\% & 91.7\% & 94.8\% \\
  75\% & 92.5\% & 91.5\% & 93.8\% \\
  90\% & 92.1\% & 93.8\% & 95.4\% \\
  99\% & 93.5\% & 90.0\% & 90.6\% \\
  \bottomrule
\end{tabular}
\caption{Ablation on truncation hyperparameter: percentage of thinking trace to keep. Double-rollout RL experiments using DAPO data on Nemotron-3-Nano-30B-A3B.}
\label{table:percentage}
\end{table}

\begin{table}[t]
\centering
\begin{tabular}{lccc}
  \toprule
  Param  & \multicolumn{3}{c}{RL with DAPO prompts} \\
  & AIME 2024 & AIME 2025 & AIME 2026 \\
  \midrule
  1  & 91.7\% & 89.2\% & 91.5\% \\
  2  & 92.7\% & 91.5\% & 94.6\% \\
  4  & 92.7\% & 91.7\% & 94.0\% \\
  8  & 92.1\% & 93.8\% & 95.4\% \\
  16 & 91.5\% & 92.9\% & 95.2\% \\
  \bottomrule
\end{tabular}
\caption{Ablation on the number of continuations per original rollout. Double-rollout RL experiments using DAPO data on Nemotron-3-Nano-30B-A3B.}
\label{table:continue}
\end{table}

Table~\ref{table:percentage} and Table~\ref{table:continue} are ablation studies on the two hyper-parameters introduced by the double-rollout method.
Table~\ref{table:percentage} presents RL experiments where we vary the percentage of thinking traces to keep before the second rollout stage.
When the percentage is below 50\%, our sampling-based reward estimation becomes noisy, -- it degenerates to difficulty assessment of the prompt in the extreme case of zero percent, -- and RL degrades the model.
When the percentage is too high at 99\%, the sampled correctness becomes nearly deterministic because the kept 99\% tokens often already include the final answer.
The results at 50\% and 75\% are not far worse than 90\%, which suggest that the proposed algorithm is robust and not overly sensitive to this hyper-parameter.
There is a clear advantage in choosing higher percentage to save compute cost of the second rollout stage and therefore 90\% is the best choice in this table.

Table~\ref{table:continue} presents RL experiments where we vary the number of continuations to sample in the second rollout stage.
This hyper-parameter is a trade-off knob between compute cost per RL step and the precision of reward estimation.
At the extreme of 1, the algorithm falls back to vanilla GRPO.
Table~\ref{table:continue} suggests an overall upward trend and that 8 is a reasonable middle ground.
It's remarkable that even two continuations would bring substantial accuracy benefit at low extra compute cost.

\begin{figure}[t]
\centering
\includegraphics[width=0.98\columnwidth]{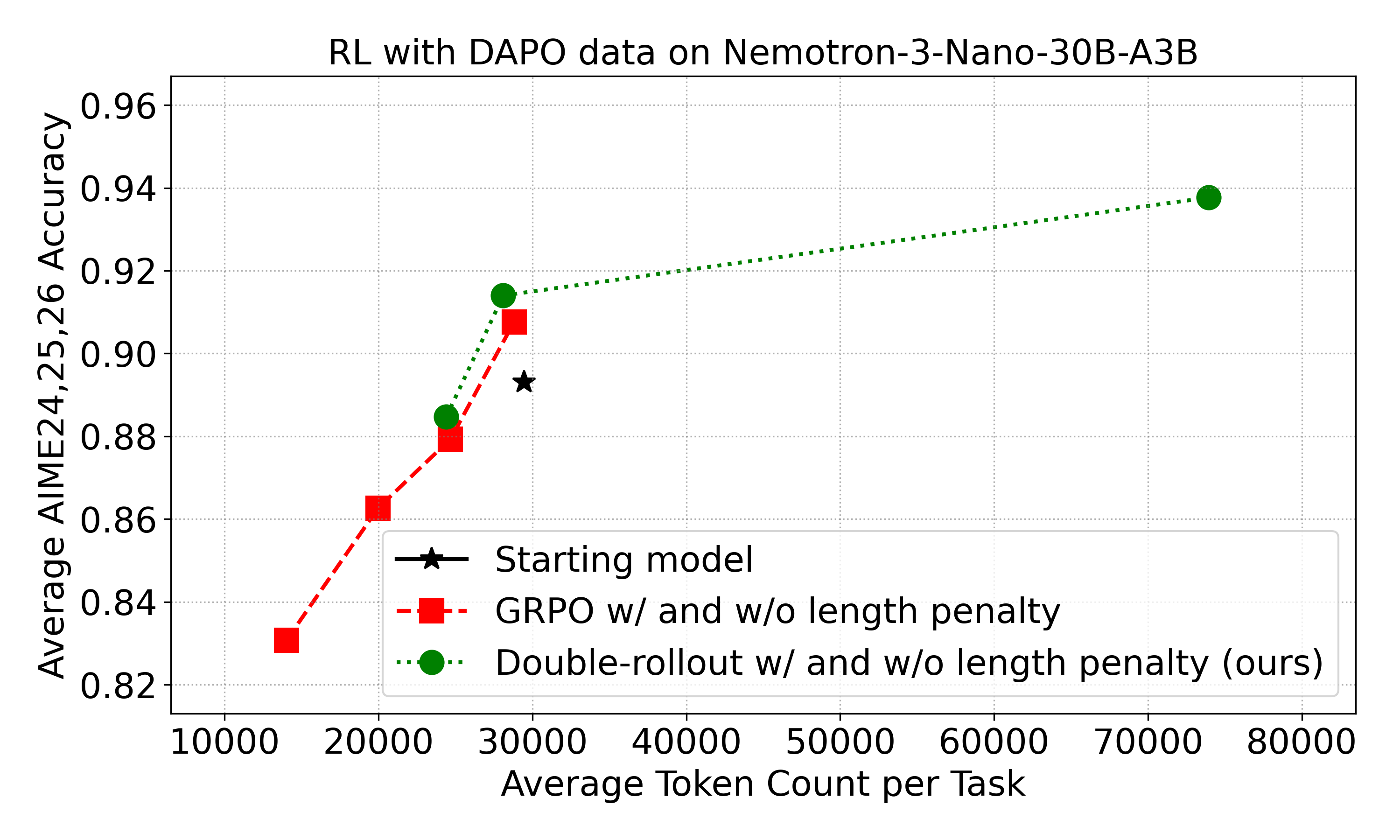}
\caption{Comparing the accuracy-efficiency Pareto frontier by the double-rollout algorithm against baselines on RL results with DAPO data on Nemotron-3-Nano-30B-A3B.}
\label{fig:tradeoff}
\end{figure}

Figure~\ref{fig:tradeoff} presents RL experiments that trade off between token efficiency and accuracy.
We use the upper-left setting in Table~\ref{table:accu} that starts from Nemotron-3-Nano-30B-A3B and performs RL with DAPO data.
We compare the two algorithms of GRPO and double-rollout, skipping GRPO with large rollout for its prohibitive compute cost.
The x axis of Figure~\ref{fig:tradeoff} is the average token usage per AIME problem and the y axis is the average of AIME 2024, AIME 2025 and AIME 2026 accuracies.
The starting model Nemotron-3-Nano-30B-A3B's trade-off point is marked with a star.
The right-most red square and right-most green circle are the same checkpoints reported in Table~\ref{table:accu} for GRPO and double-rollout respectively.
We additionally conduct two GRPO RL runs using the length penalty of \cite{arora2025training} and the two runs differ by the magnitude of length penalty.
And we conduct two more double-rollout RL runs using the adapted version of \cite{arora2025training} described at the end of Section~\ref{sec:method}, with two different magnitudes for length-based reward.
We pick Pareto-frontier checkpoints from these runs by the following criteria.
The middle green circle is the double-rollout checkpoint that generates less than 30K tokens per AIME task on average and that has the highest average accuracy across the three AIMEs.
The left green circle is the double-rollout checkpoint that generates less than 25K tokens per task and that has the highest average accuracy across the three AIMEs.
The second red square from the right is the GRPO checkpoint that generates less than 25K tokens per task and that has the highest average accuracy across the three AIMEs.
The third red square from the right is the GRPO checkpoint that generates less than 20K tokens per task and that has the highest average accuracy across the three AIMEs.
The left-most red square is the GRPO checkpoint that generates less than 15K tokens per task and that has the highest average accuracy across the three AIMEs.

Figure~\ref{fig:tradeoff} shows that the proposed double-rollout RL algorithm dominates the accuracy-efficiency Pareto frontier for the range that it covers.
It is remarkable that not a single double-rollout checkpoint uses less than 20K tokens on average per AIME task.
This holds true even when we increase the magnitude of length-based reward up to 3 per z-score: if a reliably correct rollout is shorter than average by one standard deviation, we increase its reward by 3.
The reason is that, in the proposed algorithm, a rollout must be reliably correct before it qualifies to receive the extra length-based reward (please see Section~\ref{sec:method} for details), and this prevents the model from sacrificing accuracy excessively as low-quality rollouts are never trained with positive advantage, however short and correct they may be.
This leads to a fundamentally different dynamic during RL from existing efficiency training methods.
With existing length-based penalty/reward designs, an RL run typically moves towards a succinct-but-low-accuracy state, and we either have to pick a fleeting checkpoint after a small number of steps or have to for example apply the toggle trick proposed in \cite{team2026kimi}.
With the double-rollout algorithm, an RL run goes to a steady state and then can slowly improve its accuracy-efficiency trade-off and sustain many RL steps.

If to explore the trade-off space further to the left in Figure~\ref{fig:tradeoff}, we can certainly modify the reward shaping described at the end of Section~\ref{sec:method} and achieve the effect, for example by explicitly penalizing long rollouts.
However, we hypothesize that the best approach would be to conduct long double-rollout RL experiments with length-based reward and thereby encourage the model to explore efficient-yet-reliable reasoning patterns.
Unfortunately all our experiments have been capped at 80 steps due to resource constraints.
We'll leave that to future work.

\section{Related Work}

Reinforcement learning with verifiable rewards (RLVR) with Group Relative Policy Optimization (GRPO) \cite{shao2024deepseekmath} is currently the most popular approach to train reasoning LLMs.
There have been several variants of GRPO that propose improvements on different aspects of the algorithm \cite{liu2025understanding,yu2025dapo,zheng2025group}, some of which are already included in the baseline GRPO implementation used in Section~\ref{sec:results} by NeMo-RL \cite{nemo-rl}.

Credit assignment in reasoning RL is a widely recognized challenge and has motivated many existing methods in literature that are closely related to our work.
The following are some notable examples.
VinePPO \cite{kazemnejad2024vineppo} proposes to replace the value network in PPO with Monte Carlo sampling: for each reasoning step, sequences are sampled to finish and the average reward is used to estimate the value for tokens in the step.
This works on short reasoning tasks like MATH \cite{hendrycksmath2021} and GSM8K \cite{cobbe2021gsm8k}, however the compute cost would be prohibitive for training on modern long reasoning tasks.
SPO \cite{guo2026segment} proposes a similar scheme with an additional proposal of a tree data structure to reuse some of the sampled sequences in value estimation.
However, even with all the bells and whistles, the method can only train on short reasoning tasks like MATH and GSM8K with a sequence length limit of 4K, and it is still prohibitive for training on modern long reasoning tasks.
PURE \cite{cheng2026stop} proposes to use a PRM for credit assignment and, to mitigate reward hacking due to the PRM, use the minimum of future rewards as the value function; the empirical results on math reasoning tasks, however, are fairly low accuracies.
There are numerous other works on the credit assignment problem for reasoning RL \cite{pignatelli2023survey,luo2024improve,wang2025emergent,cao2025scar}.
In contrast to these existing works, our approach strikes a practical middle ground that addresses some major failure modes of the coarse-graned GRPO and at the same time maintains practical compute cost when training on long reasoning tasks.
This is validated by our empirical results: ours is the only work on this subject that exceeds state-of-the-art math reasoning accuracy for its model size, and it outperforms across math and coding reasoning tasks and across model families.

Token efficiency is an crucial goal of RL for reasoning LLMs and many techniques have been proposed \cite{arora2025training,hou2025thinkprune,zhang2025grpo,he2025thinkdial,liu2025dler,team2026kimi}.
Most of them involve length-based reward or penalty; some enforce token budgets, which are a form of binary length-based penalty.
Some of them have reported undesirable accuracy-efficiency trade-offs.
For example, \cite{team2026kimi} reports a length-overfitting phenomenon where, when trained under token budgets, model fails to scale inference-time compute to solve hard problems; in other words, enforcing token budgets leads the model to a poor accuracy-efficiency trade-off state.
As a heuristic remedy, the authors propose to toggle the length-based penalty on/off periodically during RL.
For another example, \cite{liu2025dler} resorts to model merging in order to mitigate accuracy loss caused by RL with length penalty.
As discussed in our empirical study in Section~\ref{sec:method}, we hypothesize that the correlation between sequence lengths and
the reliability of rewards is the root cause of these undesirable trade-offs.
The proposed double-rollout RL method is able to address the issue and achieve better accuracy-efficiency trade-off points demonstrated in Figure~\ref{fig:tradeoff}.

\section{Conclusion}

This paper proposes the double-rollout reinforcement learning algorithm.
It strikes a practical middle ground between addressing the most detrimental symptoms of the credit assignment challenge in reasoning RL of LLMs and maintaining an affordable computational complexity when training on long reasoning tasks.
Empirical results are presented on training across math and coding reasoning tasks and across model families, demonstrating superior accuracies as well as superior accuracy-efficiency trade-offs.

\bibliography{aaai2027}

\begin{thebibliography}{30}
\providecommand{\natexlab}[1]{#1}

\bibitem[{Ahmad et~al.(2025)Ahmad, Narenthiran, Majumdar, Ficek, Jain, Huang, Noroozi, and Ginsburg}]{ahmad2025opencodereasoning}
Ahmad, W.~U.; Narenthiran, S.; Majumdar, S.; Ficek, A.; Jain, S.; Huang, J.; Noroozi, V.; and Ginsburg, B. 2025.
\newblock Opencodereasoning: Advancing data distillation for competitive coding.
\newblock \emph{arXiv preprint arXiv:2504.01943}.

\bibitem[{Arora and Zanette(2025)}]{arora2025training}
Arora, D.; and Zanette, A. 2025.
\newblock Training language models to reason efficiently.
\newblock \emph{arXiv preprint arXiv:2502.04463}.

\bibitem[{Blakeman et~al.(2025)Blakeman, Grattafiori, Basant, Gupta, Khattar, Renduchintala, Vavre, Shukla, Bercovich, Ficek et~al.}]{blakeman2025nemotron}
Blakeman, A.; Grattafiori, A.; Basant, A.; Gupta, A.; Khattar, A.; Renduchintala, A.; Vavre, A.; Shukla, A.; Bercovich, A.; Ficek, A.; et~al. 2025.
\newblock Nemotron 3 Nano: Open, Efficient Mixture-of-Experts Hybrid Mamba-Transformer Model for Agentic Reasoning.
\newblock \emph{arXiv preprint arXiv:2512.20848}.

\bibitem[{Blakeman et~al.(2026)Blakeman, Thomas, Jhunjhunwala, Gupta, Khattar, Rajfer, Renduchintala, Asif, Vavre, Miranda et~al.}]{blakeman2026nemotron}
Blakeman, A.; Thomas, A.; Jhunjhunwala, A.; Gupta, A.; Khattar, A.; Rajfer, A.; Renduchintala, A.; Asif, A.; Vavre, A.; Miranda, A.~F.; et~al. 2026.
\newblock Nemotron 3 Ultra: Open, Efficient Mixture-of-Experts Hybrid Mamba-Transformer Model for Agentic Reasoning.
\newblock \emph{arXiv preprint arXiv:2606.15007}.

\bibitem[{Cao et~al.(2025)Cao, Zhang, Chang, and Precup}]{cao2025scar}
Cao, M.; Zhang, S.; Chang, X.-W.; and Precup, D. 2025.
\newblock Scar: Shapley credit assignment for more efficient rlhf.
\newblock \emph{arXiv preprint arXiv:2505.20417}.

\bibitem[{Cheng et~al.(2026)Cheng, Xiong, Qiao, Li, Guo, Wang, Lv, and Wang}]{cheng2026stop}
Cheng, J.; Xiong, G.; Qiao, R.; Li, L.; Guo, C.; Wang, J.; Lv, Y.; and Wang, F.-Y. 2026.
\newblock Stop summation: Min-form credit assignment is all process reward model needs for reasoning.
\newblock \emph{Advances in Neural Information Processing Systems}, 38: 131646--131671.

\bibitem[{Cobbe et~al.(2021)Cobbe, Kosaraju, Bavarian, Chen, Jun, Kaiser, Plappert, Tworek, Hilton, Nakano, Hesse, and Schulman}]{cobbe2021gsm8k}
Cobbe, K.; Kosaraju, V.; Bavarian, M.; Chen, M.; Jun, H.; Kaiser, L.; Plappert, M.; Tworek, J.; Hilton, J.; Nakano, R.; Hesse, C.; and Schulman, J. 2021.
\newblock Training Verifiers to Solve Math Word Problems.
\newblock \emph{arXiv preprint arXiv:2110.14168}.

\bibitem[{Contributors(2025)}]{nemo-rl}
Contributors. 2025.
\newblock NeMo RL: A Scalable and Efficient Post-Training Library.
\newblock \url{https://github.com/NVIDIA-NeMo/RL}.
\newblock GitHub repository.

\bibitem[{Guo et~al.(2025)Guo, Yang, Zhang, Song, Wang, Zhu, Xu, Zhang, Ma, Bi et~al.}]{guo2025deepseek}
Guo, D.; Yang, D.; Zhang, H.; Song, J.; Wang, P.; Zhu, Q.; Xu, R.; Zhang, R.; Ma, S.; Bi, X.; et~al. 2025.
\newblock Deepseek-r1: Incentivizing reasoning capability in llms via reinforcement learning.
\newblock \emph{arXiv preprint arXiv:2501.12948}.

\bibitem[{Guo et~al.(2026)Guo, Xu, Liu, Ye, and Qiu}]{guo2026segment}
Guo, Y.; Xu, L.; Liu, J.; Ye, D.; and Qiu, S. 2026.
\newblock Segment policy optimization: Effective segment-level credit assignment in rl for large language models.
\newblock \emph{Advances in Neural Information Processing Systems}, 38: 114399--114431.

\bibitem[{He et~al.(2025)He, Yuan, Li, Wang, and Chen}]{he2025thinkdial}
He, Q.; Yuan, S.; Li, X.; Wang, M.; and Chen, J. 2025.
\newblock Thinkdial: An open recipe for controlling reasoning effort in large language models.
\newblock \emph{arXiv preprint arXiv:2508.18773}.

\bibitem[{Hendrycks et~al.(2021)Hendrycks, Burns, Kadavath, Arora, Basart, Tang, Song, and Steinhardt}]{hendrycksmath2021}
Hendrycks, D.; Burns, C.; Kadavath, S.; Arora, A.; Basart, S.; Tang, E.; Song, D.; and Steinhardt, J. 2021.
\newblock Measuring Mathematical Problem Solving With the MATH Dataset.
\newblock \emph{NeurIPS}.

\bibitem[{Hou et~al.(2025)Hou, Zhang, Ji, Liu, Qian, Andreas, and Chang}]{hou2025thinkprune}
Hou, B.; Zhang, Y.; Ji, J.; Liu, Y.; Qian, K.; Andreas, J.; and Chang, S. 2025.
\newblock Thinkprune: Pruning long chain-of-thought of llms via reinforcement learning.
\newblock \emph{arXiv preprint arXiv:2504.01296}.

\bibitem[{Jaech et~al.(2024)Jaech, Kalai, Lerer, Richardson, El-Kishky, Low, Helyar, Madry, Beutel, Carney et~al.}]{jaech2024openai}
Jaech, A.; Kalai, A.; Lerer, A.; Richardson, A.; El-Kishky, A.; Low, A.; Helyar, A.; Madry, A.; Beutel, A.; Carney, A.; et~al. 2024.
\newblock Openai o1 system card.
\newblock \emph{arXiv preprint arXiv:2412.16720}.

\bibitem[{Kazemnejad et~al.(2024)Kazemnejad, Aghajohari, Portelance, Sordoni, Reddy, Courville, and Roux}]{kazemnejad2024vineppo}
Kazemnejad, A.; Aghajohari, M.; Portelance, E.; Sordoni, A.; Reddy, S.; Courville, A.; and Roux, N.~L. 2024.
\newblock Vineppo: Refining credit assignment in rl training of llms.
\newblock \emph{arXiv preprint arXiv:2410.01679}.

\bibitem[{{Kimi Team}(2026)}]{team2026kimi}
{Kimi Team}. 2026.
\newblock Kimi K2.5: Visual Agentic Intelligence.
\newblock \emph{arXiv preprint arXiv:2602.02276}.

\bibitem[{Liu et~al.(2025{\natexlab{a}})Liu, Dong, Lu, Diao, Liu, Chen, Yin, Wang, Cheng, Choi et~al.}]{liu2025dler}
Liu, S.-Y.; Dong, X.; Lu, X.; Diao, S.; Liu, M.; Chen, M.-H.; Yin, H.; Wang, Y.-C.~F.; Cheng, K.-T.; Choi, Y.; et~al. 2025{\natexlab{a}}.
\newblock Dler: Doing length penalty right-incentivizing more intelligence per token via reinforcement learning.
\newblock \emph{arXiv preprint arXiv:2510.15110}.

\bibitem[{Liu et~al.(2025{\natexlab{b}})Liu, Chen, Li, Qi, Pang, Du, Lee, and Lin}]{liu2025understanding}
Liu, Z.; Chen, C.; Li, W.; Qi, P.; Pang, T.; Du, C.; Lee, W.~S.; and Lin, M. 2025{\natexlab{b}}.
\newblock Understanding r1-zero-like training: A critical perspective.
\newblock \emph{arXiv preprint arXiv:2503.20783}.

\bibitem[{Luo et~al.(2024)Luo, Liu, Liu, Phatale, Guo, Lara, Li, Shu, Zhu, Meng et~al.}]{luo2024improve}
Luo, L.; Liu, Y.; Liu, R.; Phatale, S.; Guo, M.; Lara, H.; Li, Y.; Shu, L.; Zhu, Y.; Meng, L.; et~al. 2024.
\newblock Improve mathematical reasoning in language models by automated process supervision.
\newblock \emph{arXiv preprint arXiv:2406.06592}.

\bibitem[{Pignatelli et~al.(2023)Pignatelli, Ferret, Geist, Mesnard, van Hasselt, Pietquin, and Toni}]{pignatelli2023survey}
Pignatelli, E.; Ferret, J.; Geist, M.; Mesnard, T.; van Hasselt, H.; Pietquin, O.; and Toni, L. 2023.
\newblock A survey of temporal credit assignment in deep reinforcement learning.
\newblock \emph{arXiv preprint arXiv:2312.01072}.

\bibitem[{{Qwen Team}(2025)}]{qwen3technicalreport}
{Qwen Team}. 2025.
\newblock Qwen3 Technical Report.
\newblock \emph{arXiv preprint arXiv:2505.09388}.

\bibitem[{{Qwen Team}(2026)}]{qwen35blog}
{Qwen Team}. 2026.
\newblock Qwen3.5: Accelerating Productivity with Native Multimodal Agents.
\newblock \url{https://qwen.ai/blog?id=qwen3.5}.

\bibitem[{Schulman et~al.(2017)Schulman, Wolski, Dhariwal, Radford, and Klimov}]{schulman2017proximal}
Schulman, J.; Wolski, F.; Dhariwal, P.; Radford, A.; and Klimov, O. 2017.
\newblock Proximal policy optimization algorithms.
\newblock \emph{arXiv preprint arXiv:1707.06347}.

\bibitem[{Shao et~al.(2024)Shao, Wang, Zhu, Xu, Song, Bi, Zhang, Zhang, Li, Wu et~al.}]{shao2024deepseekmath}
Shao, Z.; Wang, P.; Zhu, Q.; Xu, R.; Song, J.; Bi, X.; Zhang, H.; Zhang, M.; Li, Y.; Wu, Y.; et~al. 2024.
\newblock Deepseekmath: Pushing the limits of mathematical reasoning in open language models.
\newblock \emph{arXiv preprint arXiv:2402.03300}.

\bibitem[{Wang et~al.(2025)Wang, Xu, Liu, Wu, Lin, and Chen}]{wang2025emergent}
Wang, H.; Xu, Q.; Liu, C.; Wu, J.; Lin, F.; and Chen, W. 2025.
\newblock Emergent hierarchical reasoning in llms through reinforcement learning.
\newblock \emph{arXiv preprint arXiv:2509.03646}.

\bibitem[{Yu et~al.(2025)Yu, Zhang, Zhu, Yuan, Zuo, Yue, Dai, Fan, Liu, Liu et~al.}]{yu2025dapo}
Yu, Q.; Zhang, Z.; Zhu, R.; Yuan, Y.; Zuo, X.; Yue, Y.; Dai, W.; Fan, T.; Liu, G.; Liu, L.; et~al. 2025.
\newblock Dapo: An open-source llm reinforcement learning system at scale, 2025.
\newblock \emph{arXiv preprint arXiv:2503.14476}, 1: 2.

\bibitem[{Zeng et~al.(2026)Zeng, Lv, Hou, Du, Zheng, Chen, Yin, Ge, Huang, Xie et~al.}]{zeng2026glm}
Zeng, A.; Lv, X.; Hou, Z.; Du, Z.; Zheng, Q.; Chen, B.; Yin, D.; Ge, C.; Huang, C.; Xie, C.; et~al. 2026.
\newblock Glm-5: from vibe coding to agentic engineering.
\newblock \emph{arXiv preprint arXiv:2602.15763}.

\bibitem[{Zhang and Zuo(2025)}]{zhang2025grpo}
Zhang, J.; and Zuo, C. 2025.
\newblock Grpo-lead: A difficulty-aware reinforcement learning approach for concise mathematical reasoning in language models.
\newblock In \emph{Proceedings of the 2025 Conference on Empirical Methods in Natural Language Processing}, 5642--5665.

\bibitem[{Zhang et~al.(2025)Zhang, Zheng, Wu, Zhang, Lin, Yu, Liu, Zhou, and Lin}]{zhang2025lessons}
Zhang, Z.; Zheng, C.; Wu, Y.; Zhang, B.; Lin, R.; Yu, B.; Liu, D.; Zhou, J.; and Lin, J. 2025.
\newblock The lessons of developing process reward models in mathematical reasoning.
\newblock In \emph{Findings of the Association for Computational Linguistics: ACL 2025}, 10495--10516.

\bibitem[{Zheng et~al.(2025)Zheng, Liu, Li, Chen, Yu, Gao, Dang, Liu, Men, Yang et~al.}]{zheng2025group}
Zheng, C.; Liu, S.; Li, M.; Chen, X.-H.; Yu, B.; Gao, C.; Dang, K.; Liu, Y.; Men, R.; Yang, A.; et~al. 2025.
\newblock Group sequence policy optimization.
\newblock \emph{arXiv preprint arXiv:2507.18071}.

\end{thebibliography}


\end{document}